\pdfoutput=1

\documentclass[11pt]{article}

\usepackage[]{acl}
\usepackage{times}
\usepackage{latexsym}

\usepackage{amsfonts}
\usepackage{amsmath}
\usepackage{bm}
\usepackage{tabularx}
\usepackage{xurl}
\usepackage{multirow}
\usepackage{hyperref}
\usepackage{booktabs}
\usepackage{graphicx}
\usepackage{booktabs}
\usepackage{floatrow}

\usepackage{float}

\newcommand{\mlm}{\textsc{mlm}}
\newcommand{\nsp}{\textsc{nsp}}
\newcommand{\sr}{\textsc{s+r}}

\usepackage[T1]{fontenc}

\usepackage[utf8]{inputenc}

\usepackage{microtype}

\title{How does the pre-training objective affect what large language models learn about linguistic properties?}

\author{Ahmed Alajrami \and Nikolaos Aletras\\
  Department of Computer Science \\
  University of Sheffield, UK \\
  \texttt{\{ajsalajrami1, n.aletras\}@sheffield.ac.uk}
  }

\begin{document}
\maketitle
\begin{abstract}
Several pre-training objectives, such as masked language modeling (MLM), have been proposed to pre-train language models (e.g. BERT) with the aim of learning better language representations. However, to the best of our knowledge, no previous work so far has investigated how different pre-training objectives affect what BERT learns about linguistics properties. We hypothesize that linguistically motivated objectives such as MLM should help BERT to acquire better linguistic knowledge compared to other non-linguistically motivated objectives that are not intuitive or hard for humans to guess the association between the input and the label to be predicted.
To this end, we pre-train BERT with two linguistically motivated objectives and three non-linguistically motivated ones. We then probe for linguistic characteristics encoded in the representation of the resulting models.
We find strong evidence that there are only small differences in probing performance between the representations learned by the two different types of objectives. These surprising results question the dominant narrative of linguistically informed pre-training.\footnote{Code and models are available here: \url{https://github.com/aajrami/acl2022-pre-training-objectives-probing}}

\end{abstract}

\section{Introduction}

The most popular way to pre-train a transformer-based~\cite{vaswani2017attention} language model (LM), e.g. BERT \citep{devlin2019bert}, is by optimizing a masked language modeling (MLM) objective. The MLM task was inspired by the Cloze Task \citep{taylor1953cloze}, where humans were asked to guess omitted words in a sentence using its context, knowledge of syntax and other skills. The premise is that such an objective will guide a LM to encode linguistic information.

Apart from MLM, different types of objectives have been recently proposed. \citet{yang2019xlnet} introduced a pre-training objective based on token order permutations. \citet{Clark2020ELECTRA:} proposed a Replaced Token Detection pre-training task, that uses the output of a small MLM to corrupt the input by replacing some of the tokens. It then trains a discriminative model to predict if a token has been replaced or not. 
\citet{aroca-ouellette-rudzicz-2020-losses} explored various sentence and token-level auxiliary pre-training tasks (e.g. sentence ordering, term-frequency prediction), as better alternatives to the next sentence prediction (NSP) auxiliary task originally used to train BERT. 
\citet{Lan2020ALBERT:} introduced the sentence-order prediction task that focuses on the inter-sentence coherence, by predicting if two contiguous sentences have been swapped or not. \citet{iter2020pretraining} proposed another inter-sentence pre-training task, that helps LMs to encode discourse relationships between sentences using contrastive learning. \citet{yamaguchi2021frustratingly} showed that a non-linguistically intuitive task (i.e. masked first character prediction) can effectively be used for pre-training.

Meanwhile, several studies have explored how well and to what extent LMs learn linguistic information. This is usually examined using probing tasks, i.e. simple classification tasks that test the LM's encodings for a single linguistic feature such as grammatical information. It has been found through probing that BERT encodes syntactic~\citep{tenney2019you,liu2019linguistic,miaschi-dellorletta-2020-contextual,hewitt2019structural,jawahar2019does}
and semantic information~\citep{10.1162/tacl_a_00298,jawahar2019does,tenney2019you}. However, \citet{hall-maudslay-cotterell-2021-syntactic} argue that BERT's syntactic abilities may have been overestimated.

In this paper, we hypothesize that linguistically motivated objectives (e.g. MLM) should help BERT to acquire better linguistic knowledge compared to using non-linguistically motivated objectives, i.e. tasks that are hard for humans to guess the association between the input and the label to be predicted. To this end, we seek to answer the following research question: \emph{How does the pre-training objective affect what LMs learn about the English language?}

Our findings challenge the MLM status quo, showing that pre-training with non-linguistically informative objectives (\S\ref{sec:pretraing}) results in models with comparable linguistic capabilities, as measured by standard probing benchmarks (\S\ref{sec:probing}).
These surprising results (\S\ref{sec:results}) suggest that careful analysis of how LMs learn is critical to further improve language modeling (\S\ref{sec:discussion}).

\section{Pre-training Objectives}
\label{sec:pretraing}
We experiment with five different pre-training objectives. Two of them are considered linguistically motivated while the rest are not.

\subsection{Linguistically Motivated Objectives}

\paragraph{Masked Language Modeling (MLM):}
We use MLM as our first linguistically motivated pre-training objective. First introduced by \citet{devlin2019bert}, MLM randomly chooses 15\% of the tokens from the input sentence and replaces 80\% of them with a [MASK] token, 10\% with a random token, and 10\% remain unchanged. 

\paragraph{Manipulated Word Detection (S+R):}
We also experiment with a simpler linguistically motivated objective, where the model selects and replaces 10\% of input tokens with shuffled tokens from the same input sequence. Concurrently, it selects and replaces another 10\% of input tokens with random tokens from the vocabulary~\citep{yamaguchi2021frustratingly}.

\subsection{Non-Linguistically Motivated Objectives}

We assume that tasks that are hard for humans (such as a completely random prediction task) will make less likely the deeper layers of BERT (i.e. closer to the output layer) to acquire meaningful information about language. We also hypothesize that layers closer to the input might learn word co-occurrence information~\cite{sinha-etal-2021-masked}. 

\paragraph{Masked First Character Prediction (First Char):}
For our first non-linguistically motivated pre-training objective, we use the masked first character prediction introduced by \citet{yamaguchi2021frustratingly}. In this task, the model predicts only the first character of the masked token (e.g. `[c]at' and `[c]omputer' belong to the same class). The model predicts the first character as one of 29 classes, including the English alphabet and digit, punctuation mark, and other character indicators.

\paragraph{Masked ASCII Codes Summation Prediction (ASCII):}
We also propose a new non-linguistically motivated pre-training objective, where the model has to predict the summation of the ASCII code values of the characters in a masked token. To make this harder and keep the number of classes relatively small, we define a 5-way classification task by taking the modulo 5 of the ASCII summation: $V=  \left[ \sum_{i}{ascii(char_i)} \right ] \% 5$. 
Guessing the association between the input and such label, is an almost impossible task for a human.

\paragraph{Masked Random Token Classification (Random):}
Finally, we propose a completely random objective where we mask 15\% of the input tokens and we assign each masked token a class from 0 to 4 \emph{randomly} for a 5-way classification similar to the ASCII task. We assume that a model pre-trained with a random objective should not be able to learn anything meaningful about linguistic information.

\begin{table*}[ht]
\begin{center}
\small
\begin{tabular}{lcccccccc|c}
\toprule
Model & MNLI & QNLI & QQP & RTE & SST & MRPC & CoLA & STS & GLUE Avg. \\ \midrule

 & \multicolumn{9}{c}{\textsc{base} - 40 Epochs Pre-training (Upper Bound)} \\ \cmidrule{2-10}
\enskip \mlm{} + \nsp{} & 83.8 & 90.8 & 87.8 & 69.9 & 91.9 & 85.0 & 58.9 & 89.3 & 82.1 (0.4) \\ \midrule

 & \multicolumn{9}{c}{\textsc{base} - 500k Steps Pre-training} \\ \cmidrule{2-10}
\enskip \mlm{} & {\bf 81.4} & {\bf 89.0} & {\bf 86.5} & 65.1 & {\bf 90.6} & {\bf 86.0} & 52.8 & {\bf 87.2} & {\bf 79.8} $\pm$ 0.3 \\

\enskip S+R & 79.2 & 88.1 & 86.0 & {\bf 67.7} & 88.5 & 85.9 & {\bf 55.8} & {\bf 87.2} & {\bf 79.8} $\pm$ 0.3  \\

\enskip First Char & 78.8 & 87.2 & 85.4 & 60.0 & 89.1 & 83.5 & 44.5 & 85.1 & 76.7 $\pm$ 0.4  \\ 

\enskip ASCII & 76.8 & 85.3 & 84.3 & 60.8 & 87.9 & 82.2 & 42.0 & 82.4 & 75.2 $\pm$ 0.3  \\

\enskip Random &  67.5 & 63.3 & 74.9 & 53.5 & 81.7 & 71.8 & 15.1 & 23.3 & 56.4 $\pm$ 0.4 \\ \midrule

 & \multicolumn{9}{c}{\textsc{medium} - 250k Steps Pre-training} \\ \cmidrule{2-10}
\enskip \mlm{} & {\bf 78.3} & 85.6 & 85.2 & 62.2 & {\bf 90.0} & 82.0 & 44.3 & 84.0 & {\bf 76.4} $\pm$ 0.4 \\

\enskip S+R & 76.2 & 85.5 & 84.8 & {\bf 62.5} & 86.5 & 79.8 & {\bf 46.1} & {\bf 84.4} & 75.7 $\pm$ 0.1 \\

\enskip First Char & 77.7 & {\bf 85.7} & {\bf 85.4} & 58.8 & 88.7 & {\bf 82.6} & 37.4 & 83.5 & 75.0 $\pm$ 0.3  \\ 

\enskip ASCII & 75.1 & 84.4 & 83.8 & 56.6 & 87.1 & 80.5 & 34.8 & 81.2 & 72.9 $\pm$ 0.4  \\

\enskip Random & 72.9 & 81.4 & 83.1 & 54.7 & 84.0 & 73.7 & 27.3 & 76.9 & 69.3 $\pm$ 0.5  \\ \midrule

& \multicolumn{9}{c}{\textsc{small} - 250k Steps Pre-training} \\ \cmidrule{2-10}
\enskip \mlm{} & {\bf 75.8} & {\bf 84.6} & 84.4 & {\bf 59.7} & {\bf 89.0} & {\bf 81.7} & {\bf 38.7} & {\bf 83.6} & {\bf 74.7} $\pm$ 0.4 \\

\enskip S+R & 75.1 & 84.2 & 84.4 & 55.8 & 85.6 & 76.0 & 36.6 & 82.5 & 72.5 $\pm$ 0.2  \\

\enskip First Char & 74.5 & 83.3 & {\bf 84.5} & 56.3 & 87.3 & 78.4 & 35.4 & 81.4 & 72.6 $\pm$ 0.4  \\ 

\enskip ASCII & 72.9 & 82.3 & 83.1 & 55.7 & 87.0 & 72.2 & 32.8 & 77.1 & 70.4 $\pm$ 0.2 \\

\enskip Random & 70.7 & 81.0 & 82.4 & 54.4 & 84.2 & 72.5 & 23.4 & 76.2 & 68.1 $\pm$ 0.6 \\

\bottomrule
\end{tabular}
\caption{Results on GLUE dev sets with standard deviations over five runs. \textbf{Bold} values denote the best performance across each GLUE task and GLUE Avg. for each model setting.} 
\label{table:glue_result}
\end{center}
\end{table*}

\section{Probing Tasks}
\label{sec:probing}

Probing tasks \citep{adi2016fine, conneau-etal-2018-cram, hupkes2018visualisation} are used to explore in what extent linguistic properties are captured by LMs. A model is normally trained, using the representations of a language model, to predict a specific linguistic property. If it achieves high accuracy, it implies that the LM encodes that linguistic property. In this work, we use nine standard probing tasks introduced by \citet{conneau-etal-2018-cram} to examine the representation output for each layer of the different LMs we pre-train following \citet{shen2020reservoir}. These tasks probe for surface, syntactic and semantic information. The dataset for each probing task contains 100k sentences for training, 10k sentences for validation and another 10k sentences for testing.\footnote{The datasets are all publicly available by \citet{conneau2018senteval}.} We train a multi-layer perceptron (MLP) classifier for each probing task using the recommended hyperparameters in the SentEval toolkit~\cite{conneau2018senteval}.

\paragraph{Surface information task:}
\textbf{SentLen} aims for correctly predicting the number of words in a sentence.

\paragraph{Syntactic information tasks:}

\textbf{TreeDepth} tests if the representations preserve information about the hierarchical structure of a sentence, by predicting the depth of its parse tree. \textbf{TopConst} predicts the top constituents of the parse tree of a sentence. \textbf{BShift} tests if two adjacent words have been inverted or not.

\paragraph{Semantic information tasks:}
\textbf{Tense} aims to predict if the main-clause verb is present or past. \textbf{SubjNum} predicts if the subject of the main clause is singular or plural. \textbf{ObjNum} tests if the direct object of the main clause is singular or plural. Semantic Odd Man Out \textbf{(SOMO)} tests if a noun or verb has been replaced with another noun or verb. \textbf{CoordInv} predicts if a sentence made of two coordinate clauses has been inverted or not.

\section{Experiments \& Results}\label{sec:results}

\subsection{Experimental Setup}
\paragraph{Models} We pre-train BERT-\textsc{base} \citep{devlin2019bert} models by replacing MLM and the next sentence prediction (NSP) objectives, with one of the linguistically or non-linguistically motivated pre-training objectives (\S\ref{sec:pretraing}). For completeness, we also pre-train two smaller model architectures, \textsc{medium} and \textsc{small} from \citep{turc2019well} as in \citet{yamaguchi2021frustratingly}. The \textsc{medium} model has eight hidden layers and eight attention heads. The \textsc{small} model has four hidden layers and eight attention heads. Both, \textsc{medium} and \textsc{small}, models have feed-forward layers of size 2048 and hidden layers of size 512. More details on hyperprameters can be found in Appendix \ref{appendix:hyperparameter_details}.

\begin{table*}[!t]
\begin{center}
\resizebox{\textwidth}{!}{
\begin{tabular}{lccccccccc}
\toprule
{\bf Model} & {\bf SentLen} & {\bf TreeDepth} & {\bf TopConst} & {\bf BShift} & {\bf Tense} & {\bf SubjNum} & {\bf ObjNum} & {\bf SOMO} & {\bf CoordInv}  \\ 

  & (Surface) & (Syntactic) & (Syntactic) & (Syntactic) & (Semantic) & (Semantic) & (Semantic) & (Semantic) & (Semantic) \\ \midrule

& \multicolumn{9}{c}{\textsc{base} - \citet{jawahar2019does}} \\ \cmidrule{2-10}
  \mlm{}+\nsp{} & 96.2 & 41.3 & 84.1 & 87.0 & 90.0 & 88.1 & 82.2 & 65.2 & 78.7 \\
  \mlm{}+\nsp{} (untrained) & 92.5 & 29.8 & 55.2 & 50.1 & 63.8 & 67.4 & 63.7 & 50.6 & 50.3 \\ \midrule

& \multicolumn{9}{c}{\textsc{base} - 500k Steps Pre-training} \\ \cmidrule{2-10}
 \mlm{} & {\bf 96.0} $\pm$ 0.2 & 41.5 $\pm$ 0.6 & 76.9 $\pm$ 0.2 & 86.5 $\pm$ 0.1 & 88.5 $\pm$ 0.7 & 87.4 $\pm$ 1.2 & 83.8 $\pm$ 0.2 & {\bf 61.7} $\pm$ 0.5 & 65.5 $\pm$ 0.3  \\

 S+R & 92.9 $\pm$ 0.4 & {\bf 45.2} $\pm$ 0.6 & {\bf 83.6} $\pm$ 0.2 & {\bf 91.3} $\pm$ 0.7 & 87.8 $\pm$ 0.4 & 88.7 $\pm$ 0.2 & 84.5 $\pm$ 0.2 & 59.6 $\pm$ 0.4 & {\bf 69.2} $\pm$ 0.3 \\ \hline

First Char & 93.7 $\pm$ 2.4 & 43.4 $\pm$ 1.2 & 81.1 $\pm$ 0.3 & 85.0 $\pm$ 0.4 & 86.0 $\pm$ 0.3 & 88.9 $\pm$ 0.1 & {\bf 86.4} $\pm$ 0.1 & 56.5 $\pm$ 0.4 & 66.5 $\pm$ 0.8 \\ 

 ASCII & 92.9 $\pm$ 0.4 & 43.3 $\pm$ 0.7 & 81.4 $\pm$ 0.4 & 82.7 $\pm$ 0.3 & {\bf 88.7} $\pm$ 0.3 & {\bf 89.1} $\pm$ 0.3 & 84.7 $\pm$ 0.5 & 54.0 $\pm$ 0.3 & 68.5 $\pm$ 0.8  \\

 Random & 95.0 $\pm$ 0.6 & 39.6 $\pm$ 0.6 & 71.4 $\pm$ 1.0 & 68.9 $\pm$ 0.4 & 72.1 $\pm$ 0.5 & 74.3 $\pm$ 0.2 & 70.3 $\pm$ 0.1 & 50.4 $\pm$ 0.3 & 63.3 $\pm$ 0.3 \\ 
\bottomrule
\end{tabular}
}
\caption{Mean accuracy with standard deviation over three runs for the best performing layer on the probing tasks using \textsc{base} models. \textbf{Bold} values denote the best performance across each probing task.} 
\label{table:base_results}
\end{center}
\end{table*}

\paragraph{Pre-training Data} All models are pre-trained on the BookCorpus \citep{7410368} and English Wikipedia from Hugging Face.\footnote{\url{https://github.com/huggingface/datasets}} The text is tokenized using Byte-Pair-Encoding \citep{sennrich-etal-2016-neural}, resulting to a total of 2.7 billion tokens.

\paragraph{Pre-training Details} Due to limited computational resources, each \textsc{base} model is pre-trained for 500k steps, while each \textsc{medium} and \textsc{small} model is pre-trained for 250k steps using 8 NVIDIA Tesla V100 (SXM2 - 32GB). We use a batch size of 32 for \textsc{base}, and 64 for \textsc{medium} and \textsc{small}. We optimize the models using Adam \citep{kingma2014adam}.

\paragraph{Fine-tuning Details} We use the General Language Understanding Evaluation (GLUE) benchmark \citep{wang2018glue} to fine-tune each model for up to 20 epochs with early stopping. For each fine-tuning task, we use five different seeds and report the average. We report matched accuracy for MNLI task, Matthews correlation for CoLA task, Spearman correlation for STS-B task, accuracy for MRPC task, F1 scores for QQP task, and accuracy for all other tasks. The WNLI task is omitted following \citet{aroca-ouellette-rudzicz-2020-losses}.

\paragraph{BERT Representations} In all of the probing tasks, we use the BERT representations of the [CLS] token at every layer as the input to the probing classifier.

\subsection{Fine-tuning Results} 
Table \ref{table:glue_result} shows the results of fine-tuning the models with all pre-training objectives on GLUE to measure their performance in downstream tasks. For the \textsc{base} model configuration, we observe that linguistically motivated objectives (e.g. MLM, S+R) achieve the best performance in downstream tasks. However, models pre-trained with non-linguistically motivated objectives (e.g. First Char, ASCII) still achieve competitive results. As expected, the model pre-trained using the Random objective obtains the lowest performance with 56.4 GLUE average score. However, its performance is still reasonable in many downstream tasks, suggesting that the model is able to learn some co-occurrence information from the input~\cite{sinha-etal-2021-masked,yamaguchi2021frustratingly}. Similar behavior can be observed for the other two model configurations, \textsc{medium} and \textsc{small}.

\begin{table*}[!t]
\begin{center}
\resizebox{\textwidth}{!}{
\begin{tabular}{lccccccccc}
\toprule
{\bf Model} & {\bf SentLen} & {\bf TreeDepth} & {\bf TopConst} & {\bf BShift} & {\bf Tense} & {\bf SubjNum} & {\bf ObjNum} & {\bf SOMO} & {\bf CoordInv}  \\ 

  & (Surface) & (Syntactic) & (Syntactic) & (Syntactic) & (Semantic) & (Semantic) & (Semantic) & (Semantic) & (Semantic) \\ \midrule

& \multicolumn{9}{c}{\textsc{medium} - 250k Steps Pre-training} \\ \cmidrule{2-10}
 \mlm{} & 92.3 $\pm$ 0.2 & 41.1 $\pm$ 0.1 & 76.9 $\pm$ 0.5 & 80.8 $\pm$ 0.1 & 85.9 $\pm$ 0.1 & 86.7 $\pm$ 0.1 & 83.7 $\pm$ 0.5 & {\bf 56.1} $\pm$ 0.6 & 63.5 $\pm$ 0.7 \\

 S+R & {\bf 94.0} $\pm$ 0.5 & {\bf 42.6} $\pm$ 0.2 & {\bf 83.0} $\pm$ 0.5 & {\bf 84.6} $\pm$ 0.3 & 85.7 $\pm$ 0.2 & {\bf 87.9} $\pm$ 0.4 & 81.9 $\pm$ 0.5 & 55.8 $\pm$ 0.3 & {\bf 66.5} $\pm$ 1.2 \\ \hline

First Char & 93.3 $\pm$ 0.3 & 40.4 $\pm$ 0.5 & 76.8 $\pm$ 0.3 & 80.3 $\pm$ 0.4 & 85.8 $\pm$ 0.5 & 86.3 $\pm$ 1.3 & 83.1 $\pm$ 0.1 & 53.8 $\pm$ 0.6 & 61.8 $\pm$ 0.3 \\ 

 ASCII & 90.4 $\pm$ 0.5 & 40.5 $\pm$ 0.6 & 79.6 $\pm$ 0.2 & 80.0 $\pm$ 0.8 & {\bf 87.8} $\pm$ 0.5 & 85.3 $\pm$ 0.3 & 83.9 $\pm$ 0.1 & 52.7 $\pm$ 0.4 & 64.7 $\pm$ 0.1  \\

 Random & 92.9 $\pm$ 0.2 & 42.4 $\pm$ 0.8 & 71.5 $\pm$ 0.9 & 74.2 $\pm$ 0.0 & 86.1 $\pm$ 0.1 & 84.3 $\pm$ 0.3 & {\bf 85.7} $\pm$ 0.3 & 51.3 $\pm$ 0.7 & 61.5 $\pm$ 0.4 \\ 
\bottomrule
\end{tabular}
}
\caption{Mean accuracy with standard deviation over three runs for the best performing layer on the probing tasks using \textsc{medium} models. \textbf{Bold} values denote the best performance across each probing task.} 
\label{table:medium_results}
\end{center}
\end{table*}

\begin{table*}[!t]
\begin{center}
\resizebox{\textwidth}{!}{
\begin{tabular}{lccccccccc}
\toprule
{\bf Model} & {\bf SentLen} & {\bf TreeDepth} & {\bf TopConst} & {\bf BShift} & {\bf Tense} & {\bf SubjNum} & {\bf ObjNum} & {\bf SOMO} & {\bf CoordInv}  \\ 

  & (Surface) & (Syntactic) & (Syntactic) & (Syntactic) & (Semantic) & (Semantic) & (Semantic) & (Semantic) & (Semantic) \\ \midrule

& \multicolumn{9}{c}{\textsc{small} - 250k Steps Pre-training} \\ \cmidrule{2-10}
 \mlm{} & 93.7 $\pm$ 0.4 & 41.6 $\pm$ 0.2 & 73.1 $\pm$ 0.2 & 78.3 $\pm$ 0.1 & 86.4 $\pm$ 0.7 & 83.5 $\pm$ 0.2 & 83.5 $\pm$ 0.1 & {\bf 55.9} $\pm$ 0.6 & {\bf 64.0} $\pm$ 0.3 \\

 S+R & {\bf 94.7} $\pm$ 0.8 & {\bf 43.3} $\pm$ 1.0 & 76.8 $\pm$ 0.6 & {\bf 82.1} $\pm$ 0.1 & {\bf 86.5} $\pm$ 0.2 & {\bf 85.6} $\pm$ 0.3 & 83.3 $\pm$ 0.5 & 54.9 $\pm$ 0.4 & 63.9 $\pm$ 0.1 \\ \hline

First Char & 90.7 $\pm$ 0.4 & 42.3 $\pm$ 0.4 & {\bf 77.5} $\pm$ 0.1 & 76.2 $\pm$ 0.2 & 86.0 $\pm$ 0.1 & 84.7 $\pm$ 0.5 & 82.9 $\pm$ 0.7 & 52.4 $\pm$ 0.3 & {\bf 64.0} $\pm$ 0.6 \\ 

 ASCII & 89.9 $\pm$ 0.3 & 41.3 $\pm$ 0.4 & 74.6 $\pm$ 0.4 & 74.6 $\pm$ 0.1 & 85.7 $\pm$ 0.4 & 84.0 $\pm$ 0.3 & {\bf 84.4} $\pm$ 0.2 & 52.3 $\pm$ 0.4 & 62.5 $\pm$ 0.1  \\

 Random & 94.1 $\pm$ 1.0 & 42.6 $\pm$ 0.5 & 75.8 $\pm$ 0.4 & 71.0 $\pm$ 0.4 & 85.5 $\pm$ 0.5 & 83.8 $\pm$ 0.3 & 81.6 $\pm$ 0.3 & 50.7 $\pm$ 0.4 & 61.7 $\pm$ 0.5 \\ 
\bottomrule
\end{tabular}
}
\caption{Mean accuracy with standard deviation over three runs for the best performing layer on the probing tasks using \textsc{small} models. \textbf{Bold} values denote the best performance across each probing task.} 
\label{table:small_results}
\end{center}
\end{table*}

\subsection{Probing Results}

Table \ref{table:base_results} presents the results of the best performing layer on the nine probing tasks using the representations from the BERT-\textsc{base} models as inputs to the MLP classifier.
Similar to the fine-tuning results, we first observe that the predictive performance of models trained on representations learned using linguistically motivated objectives (e.g. MLM, S+R) achieve the best performance in six out of the nine probing tasks. However, \emph{models trained on the representations learned using non-linguistically motivated objectives (e.g. First Char, ASCII) achieve very competitive results.}.
For example, in the TopConst probing task, the model pre-trained using MLM pre-training objective achieves the best performance of 83.6\%, while the the model pre-trained using ASCII pre-training objective achieves 81.4\%.

Similar patterns can be observed from the probing results of the other two model configurations, \textsc{medium} and \textsc{small} (see Tables \ref{table:medium_results} and \ref{table:small_results} respectively). For instance, in the SentLen probing task in table \ref{table:medium_results}, the difference between the best performing \textsc{medium} model (S+R) and the worst performing \textsc{medium} model (ASCII) is only 3.6\%. In the ObjNum probing task in table \ref{table:small_results}, the \textsc{small} model pre-trained using a non-linguistically motivated pre-training objective (ASCII) achieves 84.4\%, while the \textsc{small} models pre-trained using linguistically motivated pre-training objectives, MLM and S+R, achieve 83.5\% and 83.3\% respectively.

The full results of the probing tasks including all layers can be found in appendix \ref{appendix:probing_results}.

\section{Discussion}\label{sec:discussion}
Theoretically, LMs with non-linguistically motivated objectives would be expected to perform drastically worse than LMs pre-trained using MLM in both downstream tasks and linguistic capabilities. However, our results show that both types of LMs have surprisingly close performance (after fine-tuning on downstream tasks) and linguistic capabilities (after probing them) using the same training data, architecture and training scheme. We speculate that the pre-training data, and the size of the models have more impact on the effectiveness of LMs than the pre-training objectives. Furthermore, the comparable performance of different objectives in probing suggests that LMs mainly learn word co-occurrence information from pre-training \cite{sinha-etal-2021-masked,yamaguchi2021frustratingly} and that the objectives may have a little effect to what actually learn about linguistic properties. 

Recent studies have explored the limitations of using probing tasks to draw conclusions over a model's linguistic knowledge with some also suggesting improvements or alternative probing methods \citep{hewitt2019designing, voita2020information, elazar2021amnesic, maudslay2021syntactic}. However, our results show no substantial differences in the performance across tasks that probe for syntactic or semantic information between models that have been pre-trained using linguistically motivated objectives or non-linguistically motivated ones.

\section{Conclusions}\label{sec:conclusions}
In this work, we compared the linguistic capabilities of LMs. Surprisingly, our results show that pre-training with linguistically motivated objectives obtain comparable performance to non-linguistically motivated objectives. This suggests that the data and the size of the model could be more influential than the objectives themselves in language modeling. In future work, we plan to extend our experiments into other languages and probing tasks.

\section*{Acknowledgments}
We would like to thank Katerina Margatina and George Chrysostomou for their invaluable feedback. We also thank the anonymous reviewers for their constructive feedback. AA is supported by the Centre for Doctoral Training in Speech and Language Technologies (SLT) and their Applications funded by UK Research and Innovation grant EP/S023062/1. NA is supported by EPSRC grant EP/V055712/1, part of the European Commission CHIST-ERA programme, call 2019 XAI: Explainable Machine Learning-based Artificial Intelligence. 

\bibliography{anthology,custom}

\begin{thebibliography}{33}
\expandafter\ifx\csname natexlab\endcsname\relax\def\natexlab#1{#1}\fi

\bibitem[{Adi et~al.(2016)Adi, Kermany, Belinkov, Lavi, and
  Goldberg}]{adi2016fine}
Yossi Adi, Einat Kermany, Yonatan Belinkov, Ofer Lavi, and Yoav Goldberg. 2016.
\newblock Fine-grained analysis of sentence embeddings using auxiliary
  prediction tasks.
\newblock \emph{arXiv preprint arXiv:1608.04207}.

\bibitem[{Aroca-Ouellette and
  Rudzicz(2020)}]{aroca-ouellette-rudzicz-2020-losses}
St{\'e}phane Aroca-Ouellette and Frank Rudzicz. 2020.
\newblock \href {https://doi.org/10.18653/v1/2020.emnlp-main.403} {{O}n
  {L}osses for {M}odern {L}anguage {M}odels}.
\newblock In \emph{Proceedings of the 2020 Conference on Empirical Methods in
  Natural Language Processing (EMNLP)}, pages 4970--4981, Online. Association
  for Computational Linguistics.

\bibitem[{Clark et~al.(2020)Clark, Luong, Le, and Manning}]{Clark2020ELECTRA:}
Kevin Clark, Minh-Thang Luong, Quoc~V. Le, and Christopher~D. Manning. 2020.
\newblock \href {https://openreview.net/forum?id=r1xMH1BtvB} {Electra:
  Pre-training text encoders as discriminators rather than generators}.
\newblock In \emph{International Conference on Learning Representations}.

\bibitem[{Conneau and Kiela(2018)}]{conneau2018senteval}
Alexis Conneau and Douwe Kiela. 2018.
\newblock Senteval: An evaluation toolkit for universal sentence
  representations.
\newblock \emph{arXiv preprint arXiv:1803.05449}.

\bibitem[{Conneau et~al.(2018)Conneau, Kruszewski, Lample, Barrault, and
  Baroni}]{conneau-etal-2018-cram}
Alexis Conneau, German Kruszewski, Guillaume Lample, Lo{\"\i}c Barrault, and
  Marco Baroni. 2018.
\newblock \href {https://doi.org/10.18653/v1/P18-1198} {What you can cram into
  a single {\$}{\&}!{\#}* vector: Probing sentence embeddings for linguistic
  properties}.
\newblock In \emph{Proceedings of the 56th Annual Meeting of the Association
  for Computational Linguistics (Volume 1: Long Papers)}, pages 2126--2136,
  Melbourne, Australia. Association for Computational Linguistics.

\bibitem[{Devlin et~al.(2019)Devlin, Chang, Lee, and
  Toutanova}]{devlin2019bert}
Jacob Devlin, Ming-Wei Chang, Kenton Lee, and Kristina Toutanova. 2019.
\newblock Bert: Pre-training of deep bidirectional transformers for language
  understanding.
\newblock In \emph{Proceedings of the 2019 Conference of the North American
  Chapter of the Association for Computational Linguistics: Human Language
  Technologies, Volume 1 (Long and Short Papers)}, pages 4171--4186.

\bibitem[{Elazar et~al.(2021)Elazar, Ravfogel, Jacovi, and
  Goldberg}]{elazar2021amnesic}
Yanai Elazar, Shauli Ravfogel, Alon Jacovi, and Yoav Goldberg. 2021.
\newblock Amnesic probing: Behavioral explanation with amnesic counterfactuals.
\newblock \emph{Transactions of the Association for Computational Linguistics},
  9:160--175.

\bibitem[{Ettinger(2020)}]{10.1162/tacl_a_00298}
Allyson Ettinger. 2020.
\newblock \href {https://doi.org/10.1162/tacl_a_00298} {{What BERT Is Not:
  Lessons from a New Suite of Psycholinguistic Diagnostics for Language
  Models}}.
\newblock \emph{Transactions of the Association for Computational Linguistics},
  8:34--48.

\bibitem[{Hall~Maudslay and
  Cotterell(2021)}]{hall-maudslay-cotterell-2021-syntactic}
Rowan Hall~Maudslay and Ryan Cotterell. 2021.
\newblock \href {https://doi.org/10.18653/v1/2021.naacl-main.11} {Do syntactic
  probes probe syntax? experiments with jabberwocky probing}.
\newblock In \emph{Proceedings of the 2021 Conference of the North American
  Chapter of the Association for Computational Linguistics: Human Language
  Technologies}, pages 124--131, Online. Association for Computational
  Linguistics.

\bibitem[{Hewitt and Liang(2019)}]{hewitt2019designing}
John Hewitt and Percy Liang. 2019.
\newblock Designing and interpreting probes with control tasks.
\newblock \emph{arXiv preprint arXiv:1909.03368}.

\bibitem[{Hewitt and Manning(2019)}]{hewitt2019structural}
John Hewitt and Christopher~D Manning. 2019.
\newblock A structural probe for finding syntax in word representations.
\newblock In \emph{Proceedings of the 2019 Conference of the North American
  Chapter of the Association for Computational Linguistics: Human Language
  Technologies, Volume 1 (Long and Short Papers)}, pages 4129--4138.

\bibitem[{Hupkes et~al.(2018)Hupkes, Veldhoen, and
  Zuidema}]{hupkes2018visualisation}
Dieuwke Hupkes, Sara Veldhoen, and Willem Zuidema. 2018.
\newblock Visualisation and'diagnostic classifiers' reveal how recurrent and
  recursive neural networks process hierarchical structure.
\newblock \emph{Journal of Artificial Intelligence Research}, 61:907--926.

\bibitem[{Iter et~al.(2020)Iter, Guu, Lansing, and
  Jurafsky}]{iter2020pretraining}
Dan Iter, Kelvin Guu, Larry Lansing, and Dan Jurafsky. 2020.
\newblock Pretraining with contrastive sentence objectives improves discourse
  performance of language models.
\newblock In \emph{Proceedings of the 58th Annual Meeting of the Association
  for Computational Linguistics}, pages 4859--4870.

\bibitem[{Jawahar et~al.(2019)Jawahar, Sagot, and Seddah}]{jawahar2019does}
Ganesh Jawahar, Beno{\^\i}t Sagot, and Djam{\'e} Seddah. 2019.
\newblock What does bert learn about the structure of language?
\newblock In \emph{ACL 2019-57th Annual Meeting of the Association for
  Computational Linguistics}.

\bibitem[{Kingma and Ba(2014)}]{kingma2014adam}
Diederik~P Kingma and Jimmy Ba. 2014.
\newblock Adam: A method for stochastic optimization.
\newblock \emph{arXiv preprint arXiv:1412.6980}.

\bibitem[{Lan et~al.(2020)Lan, Chen, Goodman, Gimpel, Sharma, and
  Soricut}]{Lan2020ALBERT:}
Zhenzhong Lan, Mingda Chen, Sebastian Goodman, Kevin Gimpel, Piyush Sharma, and
  Radu Soricut. 2020.
\newblock \href {https://openreview.net/forum?id=H1eA7AEtvS} {Albert: A lite
  bert for self-supervised learning of language representations}.
\newblock In \emph{International Conference on Learning Representations}.

\bibitem[{Liu et~al.(2019)Liu, Gardner, Belinkov, Peters, and
  Smith}]{liu2019linguistic}
Nelson~F Liu, Matt Gardner, Yonatan Belinkov, Matthew~E Peters, and Noah~A
  Smith. 2019.
\newblock Linguistic knowledge and transferability of contextual
  representations.
\newblock \emph{arXiv preprint arXiv:1903.08855}.

\bibitem[{Maudslay and Cotterell(2021)}]{maudslay2021syntactic}
Rowan~Hall Maudslay and Ryan Cotterell. 2021.
\newblock Do syntactic probes probe syntax? experiments with jabberwocky
  probing.
\newblock In \emph{Proceedings of the 2021 Conference of the North American
  Chapter of the Association for Computational Linguistics: Human Language
  Technologies}, pages 124--131.

\bibitem[{Miaschi and
  Dell{'}Orletta(2020)}]{miaschi-dellorletta-2020-contextual}
Alessio Miaschi and Felice Dell{'}Orletta. 2020.
\newblock \href {https://doi.org/10.18653/v1/2020.repl4nlp-1.15} {Contextual
  and non-contextual word embeddings: an in-depth linguistic investigation}.
\newblock In \emph{Proceedings of the 5th Workshop on Representation Learning
  for NLP}, pages 110--119, Online. Association for Computational Linguistics.

\bibitem[{Paszke et~al.(2019)Paszke, Gross, Massa, Lerer, Bradbury, Chanan,
  Killeen, Lin, Gimelshein, Antiga et~al.}]{paszke2019pytorch}
Adam Paszke, Sam Gross, Francisco Massa, Adam Lerer, James Bradbury, Gregory
  Chanan, Trevor Killeen, Zeming Lin, Natalia Gimelshein, Luca Antiga, et~al.
  2019.
\newblock Pytorch: An imperative style, high-performance deep learning library.
\newblock \emph{Advances in neural information processing systems},
  32:8026--8037.

\bibitem[{Sennrich et~al.(2016)Sennrich, Haddow, and
  Birch}]{sennrich-etal-2016-neural}
Rico Sennrich, Barry Haddow, and Alexandra Birch. 2016.
\newblock \href {https://doi.org/10.18653/v1/P16-1162} {Neural machine
  translation of rare words with subword units}.
\newblock In \emph{Proceedings of the 54th Annual Meeting of the Association
  for Computational Linguistics (Volume 1: Long Papers)}, pages 1715--1725,
  Berlin, Germany. Association for Computational Linguistics.

\bibitem[{Shen et~al.(2020)Shen, Baevski, Morcos, Keutzer, Auli, and
  Kiela}]{shen2020reservoir}
Sheng Shen, Alexei Baevski, Ari~S Morcos, Kurt Keutzer, Michael Auli, and Douwe
  Kiela. 2020.
\newblock Reservoir transformers.
\newblock \emph{arXiv preprint arXiv:2012.15045}.

\bibitem[{Sinha et~al.(2021)Sinha, Jia, Hupkes, Pineau, Williams, and
  Kiela}]{sinha-etal-2021-masked}
Koustuv Sinha, Robin Jia, Dieuwke Hupkes, Joelle Pineau, Adina Williams, and
  Douwe Kiela. 2021.
\newblock \href {https://aclanthology.org/2021.emnlp-main.230} {Masked language
  modeling and the distributional hypothesis: Order word matters pre-training
  for little}.
\newblock In \emph{Proceedings of the 2021 Conference on Empirical Methods in
  Natural Language Processing}, pages 2888--2913, Online and Punta Cana,
  Dominican Republic. Association for Computational Linguistics.

\bibitem[{Taylor(1953)}]{taylor1953cloze}
Wilson~L Taylor. 1953.
\newblock “cloze procedure”: A new tool for measuring readability.
\newblock \emph{Journalism quarterly}, 30(4):415--433.

\bibitem[{Tenney et~al.(2019)Tenney, Xia, Chen, Wang, Poliak, McCoy, Kim,
  Van~Durme, Bowman, Das et~al.}]{tenney2019you}
Ian Tenney, Patrick Xia, Berlin Chen, Alex Wang, Adam Poliak, R~Thomas McCoy,
  Najoung Kim, Benjamin Van~Durme, Samuel~R Bowman, Dipanjan Das, et~al. 2019.
\newblock What do you learn from context? probing for sentence structure in
  contextualized word representations.
\newblock \emph{arXiv preprint arXiv:1905.06316}.

\bibitem[{Turc et~al.(2019)Turc, Chang, Lee, and Toutanova}]{turc2019well}
Iulia Turc, Ming-Wei Chang, Kenton Lee, and Kristina Toutanova. 2019.
\newblock Well-read students learn better: On the importance of pre-training
  compact models.
\newblock \emph{arXiv preprint arXiv:1908.08962}.

\bibitem[{Vaswani et~al.(2017)Vaswani, Shazeer, Parmar, Uszkoreit, Jones,
  Gomez, Kaiser, and Polosukhin}]{vaswani2017attention}
Ashish Vaswani, Noam Shazeer, Niki Parmar, Jakob Uszkoreit, Llion Jones,
  Aidan~N Gomez, {\L}ukasz Kaiser, and Illia Polosukhin. 2017.
\newblock Attention is all you need.
\newblock In \emph{Advances in neural information processing systems}, pages
  5998--6008.

\bibitem[{Voita and Titov(2020)}]{voita2020information}
Elena Voita and Ivan Titov. 2020.
\newblock Information-theoretic probing with minimum description length.
\newblock In \emph{Proceedings of the 2020 Conference on Empirical Methods in
  Natural Language Processing (EMNLP)}, pages 183--196.

\bibitem[{Wang et~al.(2018)Wang, Singh, Michael, Hill, Levy, and
  Bowman}]{wang2018glue}
Alex Wang, Amanpreet Singh, Julian Michael, Felix Hill, Omer Levy, and Samuel~R
  Bowman. 2018.
\newblock Glue: A multi-task benchmark and analysis platform for natural
  language understanding.
\newblock \emph{arXiv preprint arXiv:1804.07461}.

\bibitem[{Wolf et~al.(2020)Wolf, Debut, Sanh, Chaumond, Delangue, Moi, Cistac,
  Rault, Louf, Funtowicz, Davison, Shleifer, von Platen, Ma, Jernite, Plu, Xu,
  Le~Scao, Gugger, Drame, Lhoest, and Rush}]{wolf-etal-2020-transformers}
Thomas Wolf, Lysandre Debut, Victor Sanh, Julien Chaumond, Clement Delangue,
  Anthony Moi, Pierric Cistac, Tim Rault, Remi Louf, Morgan Funtowicz, Joe
  Davison, Sam Shleifer, Patrick von Platen, Clara Ma, Yacine Jernite, Julien
  Plu, Canwen Xu, Teven Le~Scao, Sylvain Gugger, Mariama Drame, Quentin Lhoest,
  and Alexander Rush. 2020.
\newblock \href {https://doi.org/10.18653/v1/2020.emnlp-demos.6} {Transformers:
  State-of-the-art natural language processing}.
\newblock In \emph{Proceedings of the 2020 Conference on Empirical Methods in
  Natural Language Processing: System Demonstrations}, pages 38--45, Online.
  Association for Computational Linguistics.

\bibitem[{Yamaguchi et~al.(2021)Yamaguchi, Chrysostomou, Margatina, and
  Aletras}]{yamaguchi2021frustratingly}
Atsuki Yamaguchi, George Chrysostomou, Katerina Margatina, and Nikolaos
  Aletras. 2021.
\newblock Frustratingly simple pretraining alternatives to masked language
  modeling.
\newblock \emph{arXiv preprint arXiv:2109.01819}.

\bibitem[{Yang et~al.(2019)Yang, Dai, Yang, Carbonell, Salakhutdinov, and
  Le}]{yang2019xlnet}
Zhilin Yang, Zihang Dai, Yiming Yang, Jaime Carbonell, Russ~R Salakhutdinov,
  and Quoc~V Le. 2019.
\newblock Xlnet: Generalized autoregressive pretraining for language
  understanding.
\newblock \emph{Advances in neural information processing systems}, 32.

\bibitem[{Zhu et~al.(2015)Zhu, Kiros, Zemel, Salakhutdinov, Urtasun, Torralba,
  and Fidler}]{7410368}
Yukun Zhu, Ryan Kiros, Rich Zemel, Ruslan Salakhutdinov, Raquel Urtasun,
  Antonio Torralba, and Sanja Fidler. 2015.
\newblock \href {https://doi.org/10.1109/ICCV.2015.11} {Aligning books and
  movies: Towards story-like visual explanations by watching movies and reading
  books}.
\newblock In \emph{2015 IEEE International Conference on Computer Vision
  (ICCV)}, pages 19--27.

\end{thebibliography}
\bibliographystyle{acl_natbib}

\appendix
\clearpage 

\section*{Appendices}

\section{Hyperparameter Details
\label{appendix:hyperparameter_details}}
We implement the models using PyTorch \citep{paszke2019pytorch} and the Transformers library \citep{wolf-etal-2020-transformers}. We use maximum 10 epochs for \textsc{base} and \textsc{medium}, and 15 epochs for \textsc{small}. We also use a learning rate of 1e-4 for MLM. 5e-5 for \textsc{base} First Char, S+R, and ASCII. 5e-6 for \textsc{base} Random. 1e-4 for \textsc{small} and \textsc{medium} First Char, ASCII and Random. We also use weight decay of 0.01, attention dropout of 0.1, 10000 warmup steps. We also use 1e-8 Adam $\epsilon$, 0.9 Adam $\beta_1$ and 0.999 Adam $\beta_2$.

\section{Results of each Probing Task
\label{appendix:probing_results}}
Tables 5 to 13 show the full results of each of the nine probing tasks for all model architectures and layers.

\begin{table*}[!t]
\begin{center}
\small
\begin{tabular}{lccccc}
\toprule
& \multicolumn{5}{c}{{\bf SentLen}} \\ \midrule
 Layer & \multicolumn{5}{c}{\bf \textsc{base} - 500k Steps Pre-training} \\

  & \mlm{} & \sr{} & First Char & ASCII & Random \\ \midrule

1 & 95.4 $\pm$ 0.2 & 92.9 $\pm$ 0.4 & 90.7 $\pm$ 0.8 & 91.5 $\pm$ 0.3 & 92.6 $\pm$ 0.5 \\

2 & 96.0 $\pm$ 0.2 & 92.9 $\pm$ 0.2 & 92.4 $\pm$ 0.4 & 91.7 $\pm$ 0.7 & 93.6 $\pm$ 0.3 \\

3 & 95.3 $\pm$ 0.2 & 91.6 $\pm$ 0.6 & 92.9 $\pm$ 0.5 & 92.4 $\pm$ 1.7 & 94.4 $\pm$ 0.4 \\

4 & 93.8 $\pm$ 1.2 & 92.2 $\pm$ 0.8 & 93.4 $\pm$ 1.3 & 92.9 $\pm$ 1.0 & 94.1 $\pm$ 0.6 \\ 

5 & 93.9 $\pm$ 0.4 & 92.1 $\pm$ 0.6 & 93.7 $\pm$ 2.4 & 92.4 $\pm$ 0.5 & 93.8 $\pm$ 0.6 \\ 

6 & 93.6 $\pm$ 0.5 & 92.4 $\pm$ 0.5 & 93.5 $\pm$ 1.7 & 92.1 $\pm$ 0.7 & 94.3 $\pm$ 0.4  \\

7 & 92.6 $\pm$ 0.5 & 92.1 $\pm$ 0.8 & 93.1 $\pm$ 0.9 & 90.7 $\pm$ 1.4 & 94.4 $\pm$ 0.6 \\ 
 
8 & 91.2 $\pm$ 0.5 & 91.7 $\pm$ 0.5 & 92.0 $\pm$ 1.6 & 89.9 $\pm$ 1.0 & 94.2 $\pm$ 1.0 \\

9 & 89.0 $\pm$ 0.3 & 91.8 $\pm$ 0.4 & 90.9 $\pm$ 0.7 & 88.5 $\pm$ 1.6 & 95.0 $\pm$ 0.6 \\ 

10 & 82.8 $\pm$ 0.7 & 91.1 $\pm$ 0.9 & 90.0 $\pm$ 0.9 & 86.7 $\pm$ 1.7 & 94.6 $\pm$ 0.1 \\ 

11 & 79.4 $\pm$ 0.7 & 91.0 $\pm$ 0.4 & 88.6 $\pm$ 0.1 & 87.8 $\pm$ 0.5 & 94.4 $\pm$ 0.2 \\

12 & 73.9 $\pm$ 0.3 & 90.1 $\pm$ 0.3 & 85.9 $\pm$ 0.1 & 86.4 $\pm$ 0.2 & 93.6 $\pm$ 0.4 \\ \midrule

Layer & \multicolumn{5}{c}{\bf \textsc{medium} - 250k Steps Pre-training} \\

  & \mlm{} & \sr{} & First Char & ASCII & Random \\ \midrule

1 & 91.8 $\pm$ 0.5 & 88.4 $\pm$ 1.1 & 87.1 $\pm$ 0.8 & 86.6 $\pm$ 0.8 & 90.0 $\pm$ 0.9  \\

2 & 92.3 $\pm$ 0.2 & 94.0 $\pm$ 0.5 & 93.3 $\pm$ 0.3 & 90.4 $\pm$ 0.5 & 92.3 $\pm$ 0.2 \\

3 & 92.1 $\pm$ 0.2 & 94.0 $\pm$ 0.7 & 92.0 $\pm$ 0.6 & 89.2 $\pm$ 0.5 & 92.9 $\pm$ 0.2 \\

4 & 91.7 $\pm$ 0.2 & 93.4 $\pm$ 0.7 & 91.4 $\pm$ 0.2 & 89.5 $\pm$ 0.5 & 92.2 $\pm$ 0.5 \\ 

5 & 90.6 $\pm$ 0.3 & 92.7 $\pm$ 0.7 & 91.0 $\pm$ 0.2 & 89.7 $\pm$ 0.4 & 91.2 $\pm$ 0.7 \\ 

6 & 89.3 $\pm$ 0.3 & 93.0 $\pm$ 0.6 & 90.1 $\pm$ 0.8 & 89.0 $\pm$ 0.5 & 88.7 $\pm$ 0.7 \\

7 & 85.6 $\pm$ 0.2 & 92.0 $\pm$ 0.9 & 89.3 $\pm$ 0.5 & 86.1 $\pm$ 0.9 & 88.4 $\pm$ 0.7 \\ 
 
8 & 70.5 $\pm$ 0.1 & 87.8 $\pm$ 1.4 & 84.9 $\pm$ 0.5 & 83.9 $\pm$ 0.5 & 83.2 $\pm$ 0.1 \\ \midrule

Layer & \multicolumn{5}{c}{\bf \textsc{small} - 250k Steps Pre-training} \\

  & \mlm{} & \sr{} & First Char & ASCII & Random \\ \midrule

1 & 92.9 $\pm$ 0.3 & 90.3 $\pm$ 1.3 & 89.8 $\pm$ 1.1 & 89.9 $\pm$ 0.3 & 94.1 $\pm$ 1.0 \\

2 & 93.7 $\pm$ 0.4 & 93.8 $\pm$ 0.4 & 90.7 $\pm$ 0.4 & 88.7 $\pm$ 0.2 & 93.3 $\pm$ 1.1 \\

3 & 91.7 $\pm$ 0.2 & 94.7 $\pm$ 0.8 & 89.7 $\pm$ 0.2 & 86.8 $\pm$ 0.5 & 90.1 $\pm$ 1.3 \\

4 & 77.2 $\pm$ 0.3 & 93.0 $\pm$ 0.5 & 84.4 $\pm$ 0.5 & 85.5 $\pm$ 0.4 & 84.7 $\pm$ 0.3 \\ 

\bottomrule
\end{tabular}
\caption{Results of the Sentence Length (SentLen) probing task for each layer of the pre-trained models.} 
\label{table:SentLen_results}
\end{center}
\end{table*}

\begin{table*}[!t]
\begin{center}
\small

\begin{tabular}{lccccc}
\toprule
& \multicolumn{5}{c}{{\bf TreeDepth}} \\ \midrule
 Layer & \multicolumn{5}{c}{\bf \textsc{base} - 500k Steps Pre-training} \\

  & \mlm{} & \sr{} & First Char & ASCII & Random \\ \midrule

1 & 40.0 $\pm$ 0.6 & 36.6 $\pm$ 0.6 & 35.7 $\pm$ 0.2 & 36.1 $\pm$ 0.5 & 33.5 $\pm$ 0.7 \\
2 & 41.2 $\pm$ 1.1 & 38.6 $\pm$ 0.9 & 37.7 $\pm$ 0.5 & 36.6 $\pm$ 0.3 & 35.9 $\pm$ 0.5 \\
3 & 41.5 $\pm$ 0.6 & 40.0 $\pm$ 0.8 & 38.9 $\pm$ 0.6 & 37.1 $\pm$ 0.4 & 36.2 $\pm$ 0.4 \\
4 & 40.3 $\pm$ 0.7 & 41.7 $\pm$ 0.6 & 39.4 $\pm$ 0.6 & 37.7 $\pm$ 0.9 & 36.9 $\pm$ 0.4 \\
5 & 40.3 $\pm$ 1.1 & 44.2 $\pm$ 0.5 & 39.3 $\pm$ 0.3 & 38.4 $\pm$ 1.2 & 36.7 $\pm$ 0.5 \\
6 & 40.9 $\pm$ 0.7 & 45.0 $\pm$ 0.3 & 40.6 $\pm$ 0.4 & 40.7 $\pm$ 0.5 & 36.5 $\pm$ 0.5 \\
7 & 40.8 $\pm$ 0.8 & 44.9 $\pm$ 0.8 & 42.1 $\pm$ 0.6 & 42.4 $\pm$ 0.6 & 37.0 $\pm$ 0.6 \\
8 & 40.0 $\pm$ 0.7 & 45.0 $\pm$ 0.7 & 43.4 $\pm$ 1.2 & 43.3 $\pm$ 0.7 & 39.0 $\pm$ 0.3 \\
9 & 38.8 $\pm$ 1.1 & 44.3 $\pm$ 0.7 & 43.2 $\pm$ 1.3 & 43.3 $\pm$ 0.7 & 39.2 $\pm$ 0.3 \\
10 & 37.4 $\pm$ 0.3 & 45.2 $\pm$ 0.6 & 43.4 $\pm$ 1.1 & 42.9 $\pm$ 0.5 & 39.3 $\pm$ 0.5 \\
11 & 38.7 $\pm$ 0.6 & 44.5 $\pm$ 0.4 & 42.9 $\pm$ 1.2 & 42.7 $\pm$ 0.5 & 39.6 $\pm$ 0.6 \\
12 & 38.3 $\pm$ 0.3 & 42.1 $\pm$ 0.7 & 41.5 $\pm$ 0.7 & 42.3 $\pm$ 0.4 & 37.9 $\pm$ 1.3 \\ \midrule
Layer & \multicolumn{5}{c}{\bf \textsc{medium} - 250k Steps Pre-training} \\

  & \mlm{} & \sr{} & First Char & ASCII & Random \\ \midrule

1 & 37.9 $\pm$ 0.2 & 37.8 $\pm$ 0.5 & 36.4 $\pm$ 0.3 & 37.4 $\pm$ 0.1 & 36.1 $\pm$ 0.5 \\
2 & 39.0 $\pm$ 0.5 & 39.0 $\pm$ 1.2 & 36.5 $\pm$ 0.4 & 38.0 $\pm$ 0.4 & 36.4 $\pm$ 0.6 \\
3 & 39.4 $\pm$ 0.2 & 40.4 $\pm$ 0.5 & 36.3 $\pm$ 0.2 & 37.7 $\pm$ 0.6 & 38.3 $\pm$ 0.6 \\
4 & 40.5 $\pm$ 0.5 & 40.3 $\pm$ 0.6 & 36.7 $\pm$ 0.3 & 38.3 $\pm$ 0.3 & 41.6 $\pm$ 0.6 \\
5 & 41.1 $\pm$ 0.1 & 41.8 $\pm$ 1.0 & 36.9 $\pm$ 0.6 & 39.1 $\pm$ 0.5 & 42.4 $\pm$ 0.8 \\
6 & 40.5 $\pm$ 0.2 & 42.6 $\pm$ 0.2 & 37.5 $\pm$ 0.7 & 40.5 $\pm$ 0.6 & 40.5 $\pm$ 1.1 \\
7 & 39.3 $\pm$ 0.2 & 42.5 $\pm$ 0.4 & 40.4 $\pm$ 0.5 & 39.1 $\pm$ 0.8 & 39.1 $\pm$ 0.5 \\
8 & 38.6 $\pm$ 0.9 & 38.5 $\pm$ 0.6 & 40.2 $\pm$ 0.2 & 40.5 $\pm$ 0.1 & 35.6 $\pm$ 0.1 \\ 
\midrule

Layer & \multicolumn{5}{c}{\bf \textsc{small} - 250k Steps Pre-training} \\

  & \mlm{} & \sr{} & First Char & ASCII & Random \\ \midrule

1 & 37.8 $\pm$ 0.3 & 39.2 $\pm$ 0.2 & 39.1 $\pm$ 0.3 & 37.5 $\pm$ 0.2 & 38.0 $\pm$ 0.2 \\
2 & 40.1 $\pm$ 0.5 & 41.9 $\pm$ 0.6 & 40.6 $\pm$ 0.7 & 37.4 $\pm$ 0.2 & 41.6 $\pm$ 0.4 \\
3 & 39.9 $\pm$ 0.9 & 41.6 $\pm$ 0.4 & 41.2 $\pm$ 0.3 & 41.3 $\pm$ 0.4 & 42.6 $\pm$ 0.5 \\
4 & 41.6 $\pm$ 0.2 & 43.3 $\pm$ 1.0 & 42.3 $\pm$ 0.4 & 40.9 $\pm$ 0.6 & 39.2 $\pm$ 0.3 \\

\bottomrule
\end{tabular}
\caption{Results of the Tree Depth (TreeDepth) probing task for each layer of the pre-trained models.} 
\label{table:TreeDepth_results}
\end{center}
\end{table*}

\begin{table*}[!t]
\begin{center}
\small
\begin{tabular}{lccccc}
\toprule
& \multicolumn{5}{c}{{\bf TopConst}} \\ \midrule
 Layer & \multicolumn{5}{c}{\bf \textsc{base} - 500k Steps Pre-training} \\

  & \mlm{} & \sr{} & First Char & ASCII & Random \\ \midrule

1 & 62.0 $\pm$ 0.3 & 70.2 $\pm$ 0.7 & 60.9 $\pm$ 0.4 & 66.7 $\pm$ 1.1 & 65.2 $\pm$ 0.2 \\
2 & 72.6 $\pm$ 0.4 & 73.7 $\pm$ 0.2 & 69.3 $\pm$ 0.2 & 67.7 $\pm$ 0.2 & 68.4 $\pm$ 0.6 \\
3 & 74.0 $\pm$ 0.5 & 79.6 $\pm$ 0.8 & 70.7 $\pm$ 0.5 & 69.2 $\pm$ 0.2 & 69.3 $\pm$ 0.1 \\
4 & 73.0 $\pm$ 0.5 & 81.4 $\pm$ 0.4 & 71.0 $\pm$ 0.1 & 70.8 $\pm$ 0.3 & 69.9 $\pm$ 0.4 \\
5 & 73.7 $\pm$ 0.5 & 83.6 $\pm$ 0.2 & 71.3 $\pm$ 0.3 & 70.6 $\pm$ 0.5 & 69.8 $\pm$ 1.1 \\
6 & 74.6 $\pm$ 0.6 & 83.1 $\pm$ 0.7 & 71.7 $\pm$ 0.5 & 75.4 $\pm$ 0.9 & 69.2 $\pm$ 0.6 \\
7 & 75.1 $\pm$ 0.7 & 82.4 $\pm$ 0.2 & 76.2 $\pm$ 0.5 & 78.4 $\pm$ 0.5 & 70.0 $\pm$ 1.1 \\
8 & 76.9 $\pm$ 0.2 & 81.6 $\pm$ 0.4 & 78.2 $\pm$ 0.3 & 78.5 $\pm$ 0.4 & 71.4 $\pm$ 1.0 \\
9 & 76.8 $\pm$ 0.4 & 81.7 $\pm$ 0.6 & 80.1 $\pm$ 0.3 & 80.4 $\pm$ 0.2 & 70.7 $\pm$ 0.6 \\
10 & 74.6 $\pm$ 0.6 & 80.6 $\pm$ 0.7 & 81.1 $\pm$ 0.3 & 81.4 $\pm$ 0.4 & 71.2 $\pm$ 1.1 \\
11 & 74.2 $\pm$ 0.1 & 79.6 $\pm$ 0.9 & 80.7 $\pm$ 0.4 & 81.3 $\pm$ 0.6 & 69.8 $\pm$ 0.6 \\
12 & 72.5 $\pm$ 0.2 & 76.5 $\pm$ 0.5 & 79.9 $\pm$ 0.2 & 81.0 $\pm$ 0.2 & 67.4 $\pm$ 0.4 \\
\midrule
Layer & \multicolumn{5}{c}{\bf \textsc{medium} - 250k Steps Pre-training} \\

  & \mlm{} & \sr{} & First Char & ASCII & Random \\ \midrule

1 & 64.9 $\pm$ 0.3 & 63.1 $\pm$ 1.7 & 67.6 $\pm$ 0.6 & 68.2 $\pm$ 0.6 & 55.3 $\pm$ 0.3 \\
2 & 72.1 $\pm$ 0.6 & 69.8 $\pm$ 0.6 & 68.7 $\pm$ 0.5 & 70.5 $\pm$ 1.2 & 61.9 $\pm$ 1.0 \\
3 & 72.1 $\pm$ 0.6 & 72.3 $\pm$ 0.8 & 68.3 $\pm$ 0.7 & 69.1 $\pm$ 1.0 & 66.0 $\pm$ 1.4 \\
4 & 72.6 $\pm$ 0.6 & 80.6 $\pm$ 0.3 & 69.1 $\pm$ 0.6 & 74.2 $\pm$ 0.6 & 69.8 $\pm$ 0.4 \\
5 & 74.8 $\pm$ 0.5 & 81.9 $\pm$ 0.6 & 69.8 $\pm$ 0.7 & 78.1 $\pm$ 0.7 & 71.5 $\pm$ 0.9 \\
6 & 75.2 $\pm$ 0.4 & 81.9 $\pm$ 0.5 & 73.2 $\pm$ 0.1 & 79.3 $\pm$ 0.6 & 69.7 $\pm$ 0.8 \\
7 & 76.9 $\pm$ 0.5 & 83.0 $\pm$ 0.5 & 75.7 $\pm$ 0.7 & 78.5 $\pm$ 0.5 & 70.7 $\pm$ 0.6 \\
8 & 72.6 $\pm$ 0.3 & 79.8 $\pm$ 0.3 & 76.8 $\pm$ 0.3 & 79.6 $\pm$ 0.2 & 62.9 $\pm$ 0.2 \\
 \midrule

Layer & \multicolumn{5}{c}{\bf \textsc{small} - 250k Steps Pre-training} \\

  & \mlm{} & \sr{} & First Char & ASCII & Random \\ \midrule

1 & 66.4 $\pm$ 0.2 & 69.2 $\pm$ 0.4 & 74.6 $\pm$ 0.3 & 66.3 $\pm$ 0.2 & 66.7 $\pm$ 1.4 \\
2 & 72.5 $\pm$ 0.4 & 73.2 $\pm$ 0.2 & 75.8 $\pm$ 0.3 & 66.0 $\pm$ 0.5 & 74.2 $\pm$ 0.3 \\
3 & 71.9 $\pm$ 0.3 & 73.8 $\pm$ 0.2 & 76.4 $\pm$ 0.6 & 72.6 $\pm$ 0.9 & 75.8 $\pm$ 0.4 \\
4 & 73.1 $\pm$ 0.2 & 76.8 $\pm$ 0.6 & 77.5 $\pm$ 0.1 & 74.6 $\pm$ 0.4 & 72.7 $\pm$ 0.1 \\

\bottomrule
\end{tabular}
\caption{Results of the Top Constituent (TopConst) probing task for each layer of the pre-trained models.} 
\label{table:TopConst_results}
\end{center}
\end{table*}

\begin{table*}[!t]
\begin{center}
\small
\begin{tabular}{lccccc}
\toprule
& \multicolumn{5}{c}{{\bf BShift}} \\ \midrule
 Layer & \multicolumn{5}{c}{\bf \textsc{base} - 500k Steps Pre-training} \\

  & \mlm{} & \sr{} & First Char & ASCII & Random \\ \midrule

1 & 50.0 $\pm$ 0.0 & 50.0 $\pm$ 0.0 & 50.0 $\pm$ 0.0 & 50.0 $\pm$ 0.0 & 50.0 $\pm$ 0.0 \\
2 & 50.0 $\pm$ 0.1 & 50.0 $\pm$ 0.0 & 50.0 $\pm$ 0.0 & 50.0 $\pm$ 0.0 & 50.0 $\pm$ 0.0 \\
3 & 56.6 $\pm$ 0.3 & 50.0 $\pm$ 0.0 & 50.0 $\pm$ 0.0 & 50.0 $\pm$ 0.0 & 50.0 $\pm$ 0.0 \\
4 & 57.9 $\pm$ 0.2 & 74.1 $\pm$ 0.3 & 50.0 $\pm$ 0.0 & 53.4 $\pm$ 0.4 & 50.0 $\pm$ 0.0 \\
5 & 59.8 $\pm$ 0.1 & 80.7 $\pm$ 0.2 & 50.0 $\pm$ 0.0 & 50.8 $\pm$ 1.4 & 50.0 $\pm$ 0.0 \\
6 & 60.0 $\pm$ 0.7 & 83.3 $\pm$ 0.4 & 50.0 $\pm$ 0.0 & 69.6 $\pm$ 1.4 & 50.0 $\pm$ 0.0 \\
7 & 64.9 $\pm$ 0.8 & 85.6 $\pm$ 0.2 & 63.5 $\pm$ 0.6 & 73.7 $\pm$ 2.8 & 60.2 $\pm$ 1.7 \\
8 & 72.0 $\pm$ 1.3 & 88.1 $\pm$ 0.1 & 74.4 $\pm$ 0.8 & 78.5 $\pm$ 1.5 & 66.9 $\pm$ 0.2 \\
9 & 81.4 $\pm$ 0.7 & 89.5 $\pm$ 0.2 & 82.4 $\pm$ 0.7 & 81.7 $\pm$ 0.8 & 67.0 $\pm$ 0.3 \\
10 & 85.6 $\pm$ 0.2 & 90.2 $\pm$ 0.3 & 84.8 $\pm$ 0.3 & 81.7 $\pm$ 1.4 & 68.4 $\pm$ 0.2 \\
11 & 86.5 $\pm$ 0.1 & 91.2 $\pm$ 0.6 & 85.0 $\pm$ 0.4 & 82.7 $\pm$ 0.3 & 68.9 $\pm$ 0.4 \\
12 & 82.3 $\pm$ 0.3 & 91.3 $\pm$ 0.7 & 83.3 $\pm$ 0.2 & 82.4 $\pm$ 0.2 & 68.4 $\pm$ 0.1 \\
  \\ \midrule
Layer & \multicolumn{5}{c}{\bf \textsc{medium} - 250k Steps Pre-training} \\

  & \mlm{} & \sr{} & First Char & ASCII & Random \\ \midrule

1 & 50.0 $\pm$ 0.0 & 50.0 $\pm$ 0.0 & 50.0 $\pm$ 0.0 & 50.0 $\pm$ 0.0 & 50.0 $\pm$ 0.0 \\
2 & 49.8 $\pm$ 0.3 & 50.0 $\pm$ 0.0 & 50.0 $\pm$ 0.0 & 50.0 $\pm$ 0.0 & 50.0 $\pm$ 0.0 \\
3 & 49.6 $\pm$ 0.4 & 50.0 $\pm$ 0.0 & 50.0 $\pm$ 0.0 & 57.9 $\pm$ 0.5 & 65.6 $\pm$ 0.7 \\
4 & 56.2 $\pm$ 0.7 & 64.9 $\pm$ 0.3 & 50.0 $\pm$ 0.0 & 58.1 $\pm$ 0.7 & 70.5 $\pm$ 0.4 \\
5 & 64.9 $\pm$ 0.3 & 76.4 $\pm$ 0.4 & 50.9 $\pm$ 1.6 & 58.9 $\pm$ 0.8 & 74.2 $\pm$ 0.0 \\
6 & 69.6 $\pm$ 0.7 & 79.6 $\pm$ 0.1 & 73.5 $\pm$ 1.3 & 67.9 $\pm$ 1.3 & 72.5 $\pm$ 1.5 \\
7 & 80.8 $\pm$ 0.1 & 82.1 $\pm$ 0.3 & 79.9 $\pm$ 0.4 & 75.1 $\pm$ 2.7 & 73.7 $\pm$ 0.1 \\
8 & 77.9 $\pm$ 0.5 & 84.6 $\pm$ 0.3 & 80.3 $\pm$ 0.4 & 80.0 $\pm$ 0.8 & 70.3 $\pm$ 0.6 \\
 \midrule

Layer & \multicolumn{5}{c}{\bf \textsc{small} - 250k Steps Pre-training} \\

  & \mlm{} & \sr{} & First Char & ASCII & Random \\ \midrule

1 & 50.0 $\pm$ 0.1 & 50.0 $\pm$ 0.0 & 50.4 $\pm$ 0.2 & 53.2 $\pm$ 0.8 & 50.7 $\pm$ 0.4 \\
2 & 49.8 $\pm$ 0.2 & 61.9 $\pm$ 0.3 & 57.7 $\pm$ 0.1 & 60.2 $\pm$ 1.2 & 60.0 $\pm$ 0.6 \\
3 & 60.8 $\pm$ 0.7 & 74.4 $\pm$ 0.0 & 65.3 $\pm$ 0.2 & 72.1 $\pm$ 0.6 & 68.7 $\pm$ 0.7 \\
4 & 78.3 $\pm$ 0.1 & 82.1 $\pm$ 0.1 & 76.2 $\pm$ 0.2 & 74.6 $\pm$ 0.1 & 71.0 $\pm$ 0.4 \\

\bottomrule
\end{tabular}
\caption{Results of the Bigram Shift (BShift) probing task for each layer of the pre-trained models.} 
\label{table:BShift_results}
\end{center}
\end{table*}

\begin{table*}[!t]
\begin{center}
\small
\begin{tabular}{lccccc}
\toprule
& \multicolumn{5}{c}{{\bf Tense}} \\ \midrule
 Layer & \multicolumn{5}{c}{\bf \textsc{base} - 500k Steps Pre-training} \\

  & \mlm{} & \sr{} & First Char & ASCII & Random \\ \midrule

1 & 79.5 $\pm$ 0.8 & 83.6 $\pm$ 0.1 & 81.3 $\pm$ 0.1 & 79.9 $\pm$ 0.8 & 67.9 $\pm$ 0.6 \\
2 & 84.0 $\pm$ 0.7 & 84.3 $\pm$ 1.0 & 82.0 $\pm$ 0.2 & 80.3 $\pm$ 0.6 & 68.9 $\pm$ 0.8 \\
3 & 83.3 $\pm$ 0.3 & 85.7 $\pm$ 0.7 & 82.7 $\pm$ 0.5 & 82.0 $\pm$ 0.8 & 69.1 $\pm$ 0.6 \\
4 & 83.7 $\pm$ 0.7 & 86.3 $\pm$ 0.7 & 83.9 $\pm$ 0.7 & 82.9 $\pm$ 0.4 & 69.0 $\pm$ 0.3 \\
5 & 85.0 $\pm$ 0.5 & 86.3 $\pm$ 0.7 & 84.3 $\pm$ 0.9 & 83.0 $\pm$ 0.4 & 68.8 $\pm$ 0.5 \\
6 & 86.2 $\pm$ 0.2 & 87.8 $\pm$ 0.4 & 84.3 $\pm$ 0.9 & 85.3 $\pm$ 0.1 & 68.9 $\pm$ 0.4 \\
7 & 87.0 $\pm$ 0.1 & 87.1 $\pm$ 0.8 & 84.7 $\pm$ 0.6 & 86.0 $\pm$ 0.5 & 69.1 $\pm$ 0.5 \\
8 & 86.4 $\pm$ 0.8 & 87.2 $\pm$ 0.4 & 86.0 $\pm$ 0.3 & 86.1 $\pm$ 0.5 & 70.9 $\pm$ 0.1 \\
9 & 85.8 $\pm$ 1.8 & 86.3 $\pm$ 0.0 & 85.9 $\pm$ 0.2 & 87.2 $\pm$ 0.2 & 71.4 $\pm$ 0.6 \\
10 & 86.5 $\pm$ 1.5 & 85.9 $\pm$ 0.6 & 85.7 $\pm$ 0.8 & 88.5 $\pm$ 0.2 & 72.1 $\pm$ 0.5 \\
11 & 88.5 $\pm$ 0.7 & 83.7 $\pm$ 0.8 & 86.0 $\pm$ 0.7 & 88.7 $\pm$ 0.3 & 72.1 $\pm$ 0.5 \\
12 & 83.9 $\pm$ 0.0 & 81.7 $\pm$ 1.7 & 85.9 $\pm$ 0.5 & 88.6 $\pm$ 0.4 & 71.0 $\pm$ 0.4 \\
 \midrule
Layer & \multicolumn{5}{c}{\bf \textsc{medium} - 250k Steps Pre-training} \\

  & \mlm{} & \sr{} & First Char & ASCII & Random \\ \midrule

1 & 85.1 $\pm$ 0.5 & 82.2 $\pm$ 0.6 & 83.2 $\pm$ 0.4 & 81.4 $\pm$ 0.2 & 79.6 $\pm$ 0.9 \\
2 & 84.1 $\pm$ 0.5 & 84.0 $\pm$ 0.3 & 82.5 $\pm$ 0.3 & 82.4 $\pm$ 0.5 & 80.0 $\pm$ 0.8 \\
3 & 84.8 $\pm$ 0.4 & 85.4 $\pm$ 0.3 & 82.7 $\pm$ 0.1 & 82.0 $\pm$ 0.5 & 82.6 $\pm$ 0.8 \\
4 & 85.6 $\pm$ 0.6 & 85.5 $\pm$ 0.6 & 82.7 $\pm$ 0.4 & 83.4 $\pm$ 0.5 & 84.6 $\pm$ 0.7 \\
5 & 85.9 $\pm$ 0.4 & 85.0 $\pm$ 0.4 & 83.7 $\pm$ 0.4 & 84.1 $\pm$ 0.8 & 86.1 $\pm$ 0.1 \\
6 & 85.7 $\pm$ 0.8 & 85.7 $\pm$ 0.2 & 84.7 $\pm$ 0.7 & 85.4 $\pm$ 0.5 & 83.9 $\pm$ 1.5 \\
7 & 85.9 $\pm$ 0.1 & 84.6 $\pm$ 0.5 & 85.8 $\pm$ 0.5 & 85.3 $\pm$ 0.5 & 84.9 $\pm$ 0.4 \\
8 & 83.9 $\pm$ 0.5 & 82.8 $\pm$ 0.4 & 85.6 $\pm$ 0.5 & 87.8 $\pm$ 0.5 & 84.6 $\pm$ 0.5 \\
 \midrule

Layer & \multicolumn{5}{c}{\bf \textsc{small} - 250k Steps Pre-training} \\

  & \mlm{} & \sr{} & First Char & ASCII & Random \\ \midrule

1 & 86.3 $\pm$ 0.4 & 84.9 $\pm$ 0.2 & 84.7 $\pm$ 0.7 & 82.7 $\pm$ 0.6 & 84.4 $\pm$ 0.3 \\
2 & 86.2 $\pm$ 0.6 & 85.6 $\pm$ 0.5 & 84.7 $\pm$ 0.8 & 82.9 $\pm$ 0.2 & 85.2 $\pm$ 0.5 \\
3 & 86.4 $\pm$ 0.7 & 86.0 $\pm$ 0.2 & 84.7 $\pm$ 0.6 & 84.5 $\pm$ 0.8 & 85.5 $\pm$ 0.5 \\
4 & 85.2 $\pm$ 0.6 & 86.5 $\pm$ 0.2 & 86.0 $\pm$ 0.1 & 85.7 $\pm$ 0.4 & 84.9 $\pm$ 0.3 \\

\bottomrule
\end{tabular}
\caption{Results of the Tense (Tense) probing task for each layer of the pre-trained models.} 
\label{table:Tense_results}
\end{center}
\end{table*}

\begin{table*}[!t]
\begin{center}
\small
\begin{tabular}{lccccc}
\toprule
& \multicolumn{5}{c}{{\bf SubjNum}} \\ \midrule
 Layer & \multicolumn{5}{c}{\bf \textsc{base} - 500k Steps Pre-training} \\

  & \mlm{} & \sr{} & First Char & ASCII & Random \\ \midrule

1 & 75.1 $\pm$ 0.5 & 75.5 $\pm$ 0.3 & 75.7 $\pm$ 0.8 & 77.0 $\pm$ 0.1 & 69.5 $\pm$ 0.2 \\
2 & 81.6 $\pm$ 0.3 & 80.2 $\pm$ 0.3 & 78.3 $\pm$ 0.3 & 78.0 $\pm$ 0.6 & 71.7 $\pm$ 0.4 \\
3 & 82.3 $\pm$ 0.3 & 85.0 $\pm$ 0.1 & 79.1 $\pm$ 0.4 & 78.7 $\pm$ 0.5 & 72.4 $\pm$ 0.3 \\
4 & 81.8 $\pm$ 0.3 & 86.2 $\pm$ 0.5 & 79.1 $\pm$ 0.6 & 79.5 $\pm$ 0.1 & 72.1 $\pm$ 0.5 \\
5 & 83.0 $\pm$ 0.3 & 88.7 $\pm$ 0.2 & 80.3 $\pm$ 0.9 & 80.5 $\pm$ 0.2 & 72.8 $\pm$ 0.1 \\
6 & 85.0 $\pm$ 0.2 & 88.2 $\pm$ 0.3 & 82.2 $\pm$ 0.5 & 84.1 $\pm$ 0.4 & 72.7 $\pm$ 0.5 \\
7 & 84.9 $\pm$ 0.6 & 87.5 $\pm$ 0.5 & 84.3 $\pm$ 0.1 & 85.5 $\pm$ 0.4 & 73.4 $\pm$ 0.6 \\
8 & 86.0 $\pm$ 0.3 & 87.0 $\pm$ 0.9 & 85.5 $\pm$ 0.2 & 86.9 $\pm$ 0.9 & 73.9 $\pm$ 0.7 \\
9 & 87.2 $\pm$ 1.0 & 87.1 $\pm$ 0.3 & 87.9 $\pm$ 0.4 & 88.9 $\pm$ 0.6 & 73.7 $\pm$ 0.4 \\
10 & 87.4 $\pm$ 1.2 & 86.5 $\pm$ 0.5 & 88.9 $\pm$ 0.1 & 89.1 $\pm$ 0.3 & 74.3 $\pm$ 0.2 \\
11 & 86.2 $\pm$ 0.2 & 86.1 $\pm$ 0.4 & 88.1 $\pm$ 0.4 & 88.8 $\pm$ 0.3 & 74.1 $\pm$ 0.1 \\
12 & 82.3 $\pm$ 0.2 & 84.3 $\pm$ 0.4 & 86.3 $\pm$ 0.4 & 88.2 $\pm$ 0.4 & 74.2 $\pm$ 0.3 \\
 \midrule
Layer & \multicolumn{5}{c}{\bf \textsc{medium} - 250k Steps Pre-training} \\

  & \mlm{} & \sr{} & First Char & ASCII & Random \\ \midrule

1 & 79.3 $\pm$ 0.7 & 77.3 $\pm$ 0.6 & 77.0 $\pm$ 0.3 & 77.2 $\pm$ 1.1 & 75.0 $\pm$ 1.1 \\
2 & 80.7 $\pm$ 0.2 & 80.0 $\pm$ 0.1 & 78.2 $\pm$ 0.6 & 80.4 $\pm$ 0.5 & 79.9 $\pm$ 0.5 \\
3 & 81.0 $\pm$ 0.4 & 83.0 $\pm$ 0.7 & 78.0 $\pm$ 0.5 & 79.6 $\pm$ 0.2 & 80.4 $\pm$ 0.5 \\
4 & 82.5 $\pm$ 0.5 & 86.9 $\pm$ 0.3 & 79.3 $\pm$ 0.6 & 81.0 $\pm$ 0.9 & 83.4 $\pm$ 0.4 \\
5 & 83.9 $\pm$ 0.3 & 87.9 $\pm$ 0.4 & 79.7 $\pm$ 0.4 & 82.5 $\pm$ 0.5 & 84.3 $\pm$ 0.3 \\
6 & 84.5 $\pm$ 0.2 & 87.5 $\pm$ 0.3 & 83.4 $\pm$ 0.3 & 84.4 $\pm$ 0.3 & 83.1 $\pm$ 1.0 \\
7 & 86.7 $\pm$ 0.1 & 87.3 $\pm$ 0.1 & 86.3 $\pm$ 1.3 & 85.1 $\pm$ 0.5 & 83.9 $\pm$ 0.2 \\
8 & 82.5 $\pm$ 0.2 & 85.3 $\pm$ 0.5 & 85.7 $\pm$ 0.2 & 85.3 $\pm$ 0.3 & 81.0 $\pm$ 0.1 \\
 \midrule

Layer & \multicolumn{5}{c}{\bf \textsc{small} - 250k Steps Pre-training} \\

  & \mlm{} & \sr{} & First Char & ASCII & Random \\ \midrule

1 & 78.0 $\pm$ 0.8 & 80.9 $\pm$ 0.3 & 81.2 $\pm$ 0.1 & 76.5 $\pm$ 0.4 & 79.3 $\pm$ 0.3 \\
2 & 82.2 $\pm$ 0.2 & 82.5 $\pm$ 0.3 & 82.1 $\pm$ 0.4 & 76.5 $\pm$ 0.5 & 82.4 $\pm$ 0.6 \\
3 & 83.5 $\pm$ 0.2 & 81.8 $\pm$ 1.1 & 82.6 $\pm$ 0.2 & 82.6 $\pm$ 0.3 & 83.8 $\pm$ 0.3 \\
4 & 83.3 $\pm$ 0.4 & 85.6 $\pm$ 0.3 & 84.7 $\pm$ 0.5 & 84.0 $\pm$ 0.3 & 81.9 $\pm$ 0.1 \\

\bottomrule
\end{tabular}
\caption{Results of the Subject Number (SubjNum) probing task for each layer of the pre-trained models.} 
\label{table:SubjNum_results}
\end{center}
\end{table*}

\begin{table*}[!t]
\begin{center}
\small
\begin{tabular}{lccccc}
\toprule
& \multicolumn{5}{c}{{\bf ObjNum}} \\ \midrule
 Layer & \multicolumn{5}{c}{\bf \textsc{base} - 500k Steps Pre-training} \\

  & \mlm{} & \sr{} & First Char & ASCII & Random \\ \midrule

1 & 75.6 $\pm$ 0.3 & 73.6 $\pm$ 0.3 & 76.5 $\pm$ 0.4 & 77.5 $\pm$ 0.3 & 64.9 $\pm$ 0.6 \\
2 & 81.1 $\pm$ 0.1 & 77.0 $\pm$ 0.1 & 77.9 $\pm$ 0.9 & 77.7 $\pm$ 1.3 & 67.5 $\pm$ 0.4 \\
3 & 80.5 $\pm$ 1.0 & 79.7 $\pm$ 0.5 & 78.5 $\pm$ 0.5 & 79.7 $\pm$ 0.7 & 68.0 $\pm$ 0.3 \\
4 & 80.3 $\pm$ 0.8 & 81.9 $\pm$ 0.5 & 78.7 $\pm$ 3.0 & 78.6 $\pm$ 0.4 & 68.1 $\pm$ 0.1 \\
5 & 80.4 $\pm$ 1.0 & 84.4 $\pm$ 1.1 & 79.2 $\pm$ 2.9 & 78.8 $\pm$ 1.1 & 68.4 $\pm$ 0.4 \\
6 & 82.0 $\pm$ 0.1 & 84.5 $\pm$ 0.2 & 81.1 $\pm$ 1.3 & 82.2 $\pm$ 1.2 & 68.4 $\pm$ 0.6 \\
7 & 82.1 $\pm$ 0.4 & 84.4 $\pm$ 0.1 & 84.0 $\pm$ 0.7 & 83.3 $\pm$ 0.8 & 69.2 $\pm$ 0.2 \\
8 & 82.1 $\pm$ 1.0 & 84.0 $\pm$ 0.9 & 84.4 $\pm$ 0.8 & 84.3 $\pm$ 1.2 & 69.4 $\pm$ 0.2 \\
9 & 82.9 $\pm$ 0.3 & 84.1 $\pm$ 0.5 & 86.4 $\pm$ 0.1 & 84.5 $\pm$ 1.4 & 69.7 $\pm$ 0.1 \\
10 & 83.8 $\pm$ 0.2 & 82.9 $\pm$ 0.5 & 86.4 $\pm$ 0.2 & 84.7 $\pm$ 0.6 & 69.9 $\pm$ 0.2 \\
11 & 83.3 $\pm$ 0.3 & 83.8 $\pm$ 0.3 & 86.0 $\pm$ 0.3 & 84.5 $\pm$ 0.2 & 70.3 $\pm$ 0.1 \\
12 & 78.5 $\pm$ 0.3 & 81.1 $\pm$ 1.7 & 83.5 $\pm$ 0.2 & 84.7 $\pm$ 0.5 & 70.2 $\pm$ 0.3 \\
 \\ \midrule
Layer & \multicolumn{5}{c}{\bf \textsc{medium} - 250k Steps Pre-training} \\

  & \mlm{} & \sr{} & First Char & ASCII & Random \\ \midrule

1 & 80.1 $\pm$ 0.3 & 76.2 $\pm$ 0.4 & 76.2 $\pm$ 0.6 & 76.0 $\pm$ 0.3 & 75.2 $\pm$ 0.1 \\
2 & 80.1 $\pm$ 0.1 & 78.4 $\pm$ 0.2 & 77.8 $\pm$ 0.7 & 78.5 $\pm$ 0.6 & 76.4 $\pm$ 0.5 \\
3 & 80.6 $\pm$ 0.0 & 80.9 $\pm$ 0.1 & 77.2 $\pm$ 0.0 & 77.7 $\pm$ 0.8 & 78.7 $\pm$ 0.3 \\
4 & 80.7 $\pm$ 0.2 & 81.0 $\pm$ 0.4 & 78.1 $\pm$ 0.1 & 77.8 $\pm$ 1.0 & 84.6 $\pm$ 0.2 \\
5 & 82.5 $\pm$ 0.3 & 81.2 $\pm$ 0.6 & 78.7 $\pm$ 0.5 & 81.5 $\pm$ 0.2 & 85.7 $\pm$ 0.3 \\
6 & 82.9 $\pm$ 0.1 & 81.9 $\pm$ 0.5 & 81.1 $\pm$ 0.3 & 82.9 $\pm$ 0.4 & 84.2 $\pm$ 0.6 \\
7 & 83.7 $\pm$ 0.5 & 80.8 $\pm$ 0.3 & 83.1 $\pm$ 0.1 & 82.6 $\pm$ 0.2 & 83.8 $\pm$ 0.0 \\
8 & 80.2 $\pm$ 0.4 & 80.3 $\pm$ 0.5 & 81.8 $\pm$ 0.3 & 83.9 $\pm$ 0.1 & 82.2 $\pm$ 0.3 \\
 \midrule

Layer & \multicolumn{5}{c}{\bf \textsc{small} - 250k Steps Pre-training} \\

  & \mlm{} & \sr{} & First Char & ASCII & Random \\ \midrule

1 & 78.2 $\pm$ 0.9 & 81.4 $\pm$ 0.2 & 77.8 $\pm$ 0.4 & 77.7 $\pm$ 0.4 & 78.2 $\pm$ 0.3 \\
2 & 82.0 $\pm$ 0.2 & 82.4 $\pm$ 0.3 & 79.7 $\pm$ 0.2 & 78.5 $\pm$ 0.4 & 79.0 $\pm$ 0.4 \\
3 & 83.5 $\pm$ 0.1 & 82.5 $\pm$ 0.4 & 80.4 $\pm$ 0.2 & 84.4 $\pm$ 0.2 & 81.6 $\pm$ 0.3 \\
4 & 80.9 $\pm$ 0.2 & 83.3 $\pm$ 0.5 & 82.9 $\pm$ 0.7 & 83.8 $\pm$ 0.2 & 79.4 $\pm$ 0.1 \\

\bottomrule
\end{tabular}
\caption{Results of the Object Number (ObjNum) probing task for each layer of the pre-trained models.} 
\label{table:ObjNum_results}
\end{center}
\end{table*}

\begin{table*}[!t]
\begin{center}
\small
\begin{tabular}{lccccc}
\toprule
& \multicolumn{5}{c}{{\bf SOMO}} \\ \midrule
 Layer & \multicolumn{5}{c}{\bf \textsc{base} - 500k Steps Pre-training} \\

  & \mlm{} & \sr{} & First Char & ASCII & Random \\ \midrule

1 & 50.0 $\pm$ 0.2 & 50.0 $\pm$ 0.2 & 50.0 $\pm$ 0.2 & 50.0 $\pm$ 0.2 & 50.0 $\pm$ 0.2 \\
2 & 52.5 $\pm$ 0.7 & 51.6 $\pm$ 0.3 & 50.5 $\pm$ 1.1 & 50.0 $\pm$ 0.2 & 50.0 $\pm$ 0.2 \\
3 & 54.4 $\pm$ 1.3 & 50.0 $\pm$ 0.2 & 51.8 $\pm$ 0.9 & 50.0 $\pm$ 0.2 & 50.0 $\pm$ 0.2 \\
4 & 55.2 $\pm$ 0.5 & 53.7 $\pm$ 0.8 & 52.5 $\pm$ 0.5 & 50.7 $\pm$ 1.2 & 50.0 $\pm$ 0.2 \\
5 & 55.8 $\pm$ 0.0 & 55.4 $\pm$ 0.1 & 52.1 $\pm$ 0.8 & 50.0 $\pm$ 0.2 & 50.0 $\pm$ 0.2 \\
6 & 57.6 $\pm$ 0.7 & 56.1 $\pm$ 0.3 & 52.8 $\pm$ 0.2 & 50.0 $\pm$ 0.2 & 50.0 $\pm$ 0.2 \\
7 & 58.2 $\pm$ 1.0 & 56.8 $\pm$ 0.5 & 52.8 $\pm$ 1.1 & 50.0 $\pm$ 0.2 & 50.0 $\pm$ 0.2 \\
8 & 58.1 $\pm$ 0.6 & 56.9 $\pm$ 1.3 & 53.7 $\pm$ 0.7 & 50.0 $\pm$ 0.2 & 50.0 $\pm$ 0.2 \\
9 & 59.1 $\pm$ 0.4 & 57.9 $\pm$ 1.5 & 54.1 $\pm$ 1.0 & 53.2 $\pm$ 0.9 & 50.0 $\pm$ 0.2 \\
10 & 60.6 $\pm$ 0.5 & 58.5 $\pm$ 0.9 & 56.3 $\pm$ 0.7 & 53.4 $\pm$ 0.2 & 50.4 $\pm$ 0.3 \\
11 & 61.7 $\pm$ 0.5 & 58.9 $\pm$ 0.6 & 56.5 $\pm$ 0.4 & 53.9 $\pm$ 1.0 & 50.2 $\pm$ 0.3 \\
12 & 57.8 $\pm$ 0.4 & 59.6 $\pm$ 0.4 & 55.4 $\pm$ 1.0 & 54.0 $\pm$ 0.3 & 50.2 $\pm$ 0.5 \\
 \\ \midrule
Layer & \multicolumn{5}{c}{\bf \textsc{medium} - 250k Steps Pre-training} \\

  & \mlm{} & \sr{} & First Char & ASCII & Random \\ \midrule

1 & 51.6 $\pm$ 0.5 & 50.2 $\pm$ 0.3 & 50.0 $\pm$ 0.2 & 50.7 $\pm$ 0.8 & 50.0 $\pm$ 0.2 \\
2 & 52.3 $\pm$ 0.7 & 51.1 $\pm$ 0.1 & 50.0 $\pm$ 0.2 & 52.2 $\pm$ 0.4 & 50.0 $\pm$ 0.2 \\
3 & 53.2 $\pm$ 0.1 & 52.6 $\pm$ 0.4 & 50.0 $\pm$ 0.2 & 52.1 $\pm$ 0.3 & 50.0 $\pm$ 0.2 \\
4 & 53.1 $\pm$ 0.8 & 52.9 $\pm$ 0.7 & 50.0 $\pm$ 0.2 & 51.3 $\pm$ 0.3 & 50.8 $\pm$ 0.3 \\
5 & 53.5 $\pm$ 0.6 & 53.8 $\pm$ 0.6 & 50.0 $\pm$ 0.2 & 51.2 $\pm$ 0.4 & 51.0 $\pm$ 0.4 \\
6 & 54.6 $\pm$ 1.1 & 53.9 $\pm$ 0.7 & 51.5 $\pm$ 1.5 & 51.3 $\pm$ 0.2 & 50.0 $\pm$ 0.1 \\
7 & 56.1 $\pm$ 0.6 & 55.2 $\pm$ 0.6 & 53.2 $\pm$ 0.2 & 52.0 $\pm$ 0.2 & 51.3 $\pm$ 0.7 \\
8 & 54.1 $\pm$ 0.1 & 55.8 $\pm$ 0.3 & 53.8 $\pm$ 0.6 & 52.7 $\pm$ 0.4 & 50.6 $\pm$ 0.3 \\
 \midrule

Layer & \multicolumn{5}{c}{\bf \textsc{small} - 250k Steps Pre-training} \\

  & \mlm{} & \sr{} & First Char & ASCII & Random \\ \midrule

1 & 52.5 $\pm$ 0.2 & 52.2 $\pm$ 0.2 & 51.8 $\pm$ 0.2 & 51.3 $\pm$ 0.3 & 50.4 $\pm$ 0.3 \\
2 & 55.4 $\pm$ 0.2 & 54.3 $\pm$ 0.4 & 51.5 $\pm$ 0.2 & 51.1 $\pm$ 0.2 & 50.7 $\pm$ 0.4 \\
3 & 55.9 $\pm$ 0.6 & 54.8 $\pm$ 0.8 & 51.5 $\pm$ 0.2 & 52.2 $\pm$ 0.0 & 50.6 $\pm$ 0.2 \\
4 & 53.9 $\pm$ 0.7 & 54.9 $\pm$ 0.4 & 52.4 $\pm$ 0.3 & 52.3 $\pm$ 0.4 & 50.2 $\pm$ 0.5 \\

\bottomrule
\end{tabular}
\caption{Results of the Semantic Odd Man Out (SOMO) probing task for each layer of the pre-trained models.} 
\label{table:SOMO_results}
\end{center}
\end{table*}

\begin{table*}[!t]
\begin{center}
\small
\begin{tabular}{lccccc}
\toprule
& \multicolumn{5}{c}{{\bf CoordInv}} \\ \midrule
 Layer & \multicolumn{5}{c}{\bf \textsc{base} - 500k Steps Pre-training} \\

  & \mlm{} & \sr{} & First Char & ASCII & Random \\ \midrule

1 & 57.3 $\pm$ 1.1 & 56.5 $\pm$ 1.0 & 55.1 $\pm$ 1.6 & 50.0 $\pm$ 0.0 & 50.0 $\pm$ 0.0 \\
2 & 61.0 $\pm$ 0.5 & 59.7 $\pm$ 0.5 & 58.0 $\pm$ 0.6 & 50.0 $\pm$ 0.0 & 51.7 $\pm$ 3.0 \\
3 & 61.8 $\pm$ 0.8 & 63.5 $\pm$ 0.8 & 58.9 $\pm$ 0.3 & 57.2 $\pm$ 0.6 & 57.8 $\pm$ 0.3 \\
4 & 61.2 $\pm$ 0.5 & 64.8 $\pm$ 1.4 & 59.4 $\pm$ 0.6 & 59.6 $\pm$ 0.5 & 52.3 $\pm$ 4.0 \\
5 & 62.0 $\pm$ 0.6 & 67.6 $\pm$ 0.4 & 60.2 $\pm$ 0.7 & 59.1 $\pm$ 0.3 & 55.2 $\pm$ 4.5 \\
6 & 62.8 $\pm$ 0.4 & 69.2 $\pm$ 0.3 & 59.6 $\pm$ 0.7 & 59.8 $\pm$ 1.6 & 58.2 $\pm$ 0.4 \\
7 & 61.6 $\pm$ 0.6 & 68.0 $\pm$ 0.3 & 61.3 $\pm$ 0.9 & 61.5 $\pm$ 2.0 & 59.8 $\pm$ 0.2 \\
8 & 62.1 $\pm$ 0.4 & 67.4 $\pm$ 0.4 & 63.4 $\pm$ 0.7 & 62.9 $\pm$ 2.1 & 61.4 $\pm$ 0.2 \\
9 & 62.1 $\pm$ 1.0 & 66.9 $\pm$ 0.2 & 63.9 $\pm$ 0.9 & 66.0 $\pm$ 1.0 & 62.6 $\pm$ 1.0 \\
10 & 64.4 $\pm$ 0.5 & 67.8 $\pm$ 0.2 & 65.6 $\pm$ 0.6 & 67.6 $\pm$ 1.1 & 63.0 $\pm$ 0.2 \\
11 & 65.5 $\pm$ 0.3 & 67.7 $\pm$ 0.5 & 66.5 $\pm$ 0.8 & 68.4 $\pm$ 0.5 & 63.3 $\pm$ 0.3 \\
12 & 63.7 $\pm$ 1.3 & 65.4 $\pm$ 0.4 & 64.4 $\pm$ 0.9 & 68.5 $\pm$ 0.8 & 61.3 $\pm$ 0.7 \\
\\ \midrule
Layer & \multicolumn{5}{c}{\bf \textsc{medium} - 250k Steps Pre-training} \\

  & \mlm{} & \sr{} & First Char & ASCII & Random \\ \midrule

1 & 59.4 $\pm$ 0.2 & 57.7 $\pm$ 0.3 & 56.9 $\pm$ 0.8 & 56.7 $\pm$ 0.7 & 55.9 $\pm$ 1.6 \\
2 & 63.5 $\pm$ 0.7 & 60.7 $\pm$ 0.8 & 56.7 $\pm$ 0.4 & 60.4 $\pm$ 0.5 & 57.9 $\pm$ 0.2 \\
3 & 62.1 $\pm$ 0.0 & 63.6 $\pm$ 0.4 & 56.5 $\pm$ 0.1 & 59.3 $\pm$ 0.7 & 58.6 $\pm$ 0.2 \\
4 & 62.5 $\pm$ 0.2 & 65.6 $\pm$ 1.0 & 56.0 $\pm$ 0.7 & 60.0 $\pm$ 0.7 & 61.5 $\pm$ 0.4 \\
5 & 63.1 $\pm$ 0.3 & 66.2 $\pm$ 1.2 & 57.6 $\pm$ 1.2 & 60.2 $\pm$ 0.4 & 61.4 $\pm$ 0.5 \\
6 & 62.5 $\pm$ 0.3 & 65.7 $\pm$ 1.5 & 58.3 $\pm$ 0.4 & 60.2 $\pm$ 1.0 & 60.1 $\pm$ 0.7 \\
7 & 61.7 $\pm$ 0.6 & 66.5 $\pm$ 1.2 & 60.4 $\pm$ 0.9 & 60.1 $\pm$ 1.5 & 60.3 $\pm$ 0.7 \\
8 & 58.4 $\pm$ 0.5 & 63.8 $\pm$ 1.8 & 61.8 $\pm$ 0.3 & 64.7 $\pm$ 0.1 & 58.7 $\pm$ 0.4 \\
 \midrule

Layer & \multicolumn{5}{c}{\bf \textsc{small} - 250k Steps Pre-training} \\

  & \mlm{} & \sr{} & First Char & ASCII & Random \\ \midrule

1 & 61.4 $\pm$ 0.1 & 60.1 $\pm$ 0.6 & 62.8 $\pm$ 0.1 & 59.7 $\pm$ 0.4 & 58.9 $\pm$ 0.5 \\
2 & 64.0 $\pm$ 0.3 & 62.2 $\pm$ 0.4 & 64.0 $\pm$ 0.6 & 59.1 $\pm$ 0.2 & 61.0 $\pm$ 0.8 \\
3 & 62.2 $\pm$ 0.3 & 62.7 $\pm$ 0.3 & 63.0 $\pm$ 0.4 & 61.4 $\pm$ 0.2 & 61.7 $\pm$ 0.5 \\
4 & 59.4 $\pm$ 0.5 & 63.9 $\pm$ 0.1 & 62.2 $\pm$ 0.3 & 62.5 $\pm$ 0.1 & 59.9 $\pm$ 0.2 \\

\bottomrule
\end{tabular}
\caption{Results of the Coordination Inversion (CoordInv) probing task for each layer of the pre-trained models.} 
\label{table:CoordInv_results}
\end{center}
\end{table*}

\end{document}